\documentclass[sigconf]{acmart}

\AtBeginDocument{%
  }

\setcopyright{rightsretained}
\copyrightyear{2023}
\acmYear{2023}
\acmDOI{10.1145/3581783.3612587}

\acmConference[MM '23]{Proceedings of the 31st ACM International Conference on Multimedia}{October 29-November 3, 2023}{Ottawa, ON, Canada}
\acmBooktitle{Proceedings of the 31st ACM International Conference on Multimedia (MM '23), October 29-November 3, 2023, Ottawa, ON, Canada}

\acmPrice{15.00}
\acmISBN{979-8-4007-0108-5/23/10}

\newcommand{\Sref}[1]{Sec.~\ref{#1}}
\newcommand{\Fref}[1]{Fig.~\ref{#1}}
\newcommand{\Tref}[1]{Table~\ref{#1}}
\newcommand{\Eref}[1]{Eqn.~\ref{#1}}
\newcommand{\cmark}{\text{\ding{51}}}
\newcommand{\xmark}{\text{\ding{55}}}

\usepackage{tabularray}
\usepackage{wasysym}
\usepackage{multirow}
\usepackage{graphicx}
\usepackage{amsmath}
\usepackage{pifont}

\begin{document}

\title{That's What I Said:\\ Fully-Controllable Talking Face Generation}


\author{Youngjoon Jang$^*$}
\email{wgs01088@kaist.ac.kr}
\affiliation{%
  \institution{KAIST}
  \city{Daejeon}
  \country{Republic of Korea}}

\author{Kyeongha Rho$^*$}
\email{khrho325@kaist.ac.kr}
\affiliation{%
  \institution{KAIST}
  \city{Daejeon}
  \country{Republic of Korea}}

\author{Jongbhin Woo}
\email{jongbin.woo@kaist.ac.kr}
\affiliation{%
  \institution{KAIST}
  \city{Daejeon}
  \country{Republic of Korea}}

\author{Hyeongkeun Lee}
\email{lhk528@kaist.ac.kr}
\affiliation{%
  \institution{KAIST}
  \city{Daejeon}
  \country{Republic of Korea}}

\author{Jihwan Park}
\email{jihwan.park@42dot.ai}
\affiliation{%
  \institution{42dot Inc., Hyundai Motor Company}
  \city{Seoul}
  \country{Republic of Korea}}

\author{Youshin Lim}
\email{youshin.lim@42dot.ai}
\affiliation{%
  \institution{42dot Inc., Hyundai Motor Company}
  \city{Seoul}
  \country{Republic of Korea}}

\author{Byeong-Yeol Kim}
\email{byeongyeol.kim@42dot.ai}
\affiliation{%
  \institution{42dot Inc., Hyundai Motor Company}
  \city{Seoul}
  \country{Republic of Korea}}

\author{Joon Son Chung}
\email{joonson@kaist.ac.kr}
\affiliation{%
  \institution{KAIST}
  \city{Daejeon}
  \country{Republic of Korea}}

\thanks{* These authors contributed equally.}
\renewcommand{\shortauthors}{Youngjoon Jang et al.}


\begin{abstract}
    The goal of this paper is to synthesise talking faces with controllable facial motions.
    To achieve this goal, we propose two key ideas. 
    The first is to establish a canonical space where every face has the same motion patterns but different identities.
    The second is to navigate a multimodal motion space that only represents motion-related features while eliminating identity information. 
    To disentangle identity and motion, we introduce an orthogonality constraint between the two different latent spaces.
    From this, our method can generate natural-looking talking faces with fully controllable facial attributes and accurate lip synchronisation.
    Extensive experiments demonstrate that our method achieves state-of-the-art results in terms of both visual quality and lip-sync score. 
    To the best of our knowledge, we are the first to develop a talking face generation framework that can accurately manifest full target facial motions including lip, head pose, and eye movements in the generated video without any additional supervision beyond RGB video with audio.
\end{abstract}

\ccsdesc[500]{Information systems~Multimedia content creation}
\ccsdesc[300]{Computing methodologies~Image processing}

\keywords{Talking Face Generation, Face Animation, Motion Transfer}
\begin{teaserfigure}
\centering
  \includegraphics[width=0.90\textwidth]{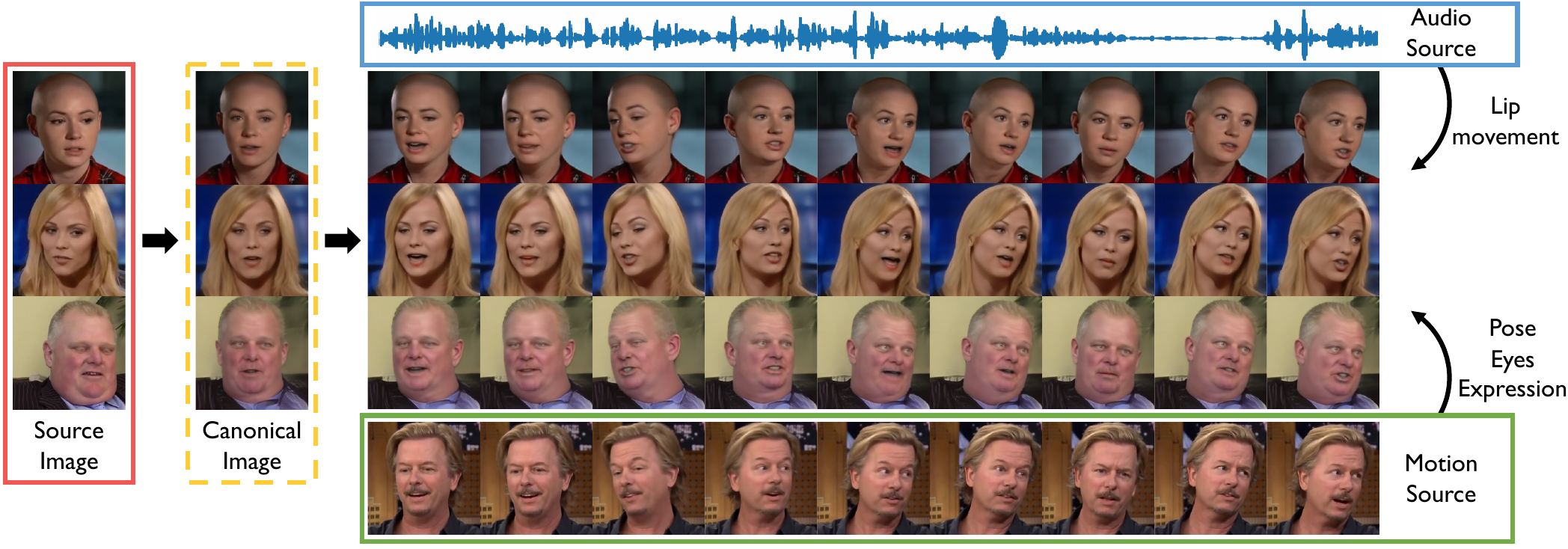}
  \vspace{-2mm}
  \caption{Our Fully-Controllable Talking Face Generation (FC-TFG) framework precisely reflects every facial expression of the motion source while synchronising the lip shape with the input audio source. 
    The key to our framework is to find the canonical space, where every face has the same motion patterns, but has different identities.}
  \Description{.}
  \label{fig:teaser}
\end{teaserfigure}

\maketitle

\section{Introduction}
Audio-driven talking face generation technology has numerous applications in the film and entertainment industry, including virtual assistants, video conferencing, and dubbing. Its primary goal is to generate animated faces that closely match the audio, creating more engaging and interactive experiences for users. This technology has the potential to revolutionise various industries by making human-machine interactions more natural and immersive.

Previous literature on deep learning-based talking face synthesis can be divided into two branches.
The first branch ~\cite{chung2017you,zhu2018arbitrary,chen2018lip,zhou2019talking,vougioukas2020realistic,song2018talking,prajwal2020lip,park2022synctalkface} only uses RGB modality as a form of supervision to reconstruct a target video, while the second branch~\cite{suwajanakorn2017synthesizing,das2020speech,chen2019hierarchical,anderson2013expressive,thies2020neural,song2022everybody,chen2020talking,zhou2020makelttalk,richard2021audio,ji2021audio,wang2022one} leverages 2D or 3D structural information to propagate more detailed supervision. These works have made significant progress in generating natural-looking lip motions. Nevertheless, there is still a need for further development in controlling head pose and refining facial attributes in greater detail.
Some methods~\cite{zhou2021pose, liang2022expressive, min2022styletalker, wang2022progressive, hwang2023discohead} have the ability to generate talking faces that closely resemble the movements and identities of a target video. 
However, these methods still have limitations that hinder their practical application. 
For instance, \cite{zhou2021pose} is only capable of altering the head pose, while \cite{min2022styletalker} is incapable of creating detailed facial components like eye gaze movements. \cite{liang2022expressive, wang2022progressive, hwang2023discohead} require the utilisation of facial keypoints to physically separate facial components such as eyes and lips to create varied expressions. 

In this paper, we propose a novel framework, Fully-Controllable Talking Face Generation (FC-TFG), which aims to generate talking faces that exactly copy full target motion including head pose, eyebrows, eye blinks, and eye gaze movements. 
The key to our method is preserving semantically meaningful features associated with the identity of each person via navigation of the latent space without requiring additional supervision such as facial keypoints. 

Our work is raised from the fundamental question in talking face generation: is it possible to completely disentangle facial motions and facial identities in the latent space?
Recent works~\cite{shen2020interpret, jahanian2020steering, goetschalckx2019ganalyze, voynov2020ganlatent} discover that semantically meaningful directions exist in the latent space of Generative Adversarial Networks (GANs) by generating face images with target facial attributes.
Based on these findings, \cite{wang2022latent} proposes a method that uses pre-defined orthogonal vectors representing basic visual transformations needed for face animations. However, using the motion dictionary fails to manipulate every detail of facial attributes such as eye gaze movements and lip shape due to the limited granularity of motion vectors.

\begin{figure*}[!t]
    \centering
    \includegraphics[width=0.93\linewidth]{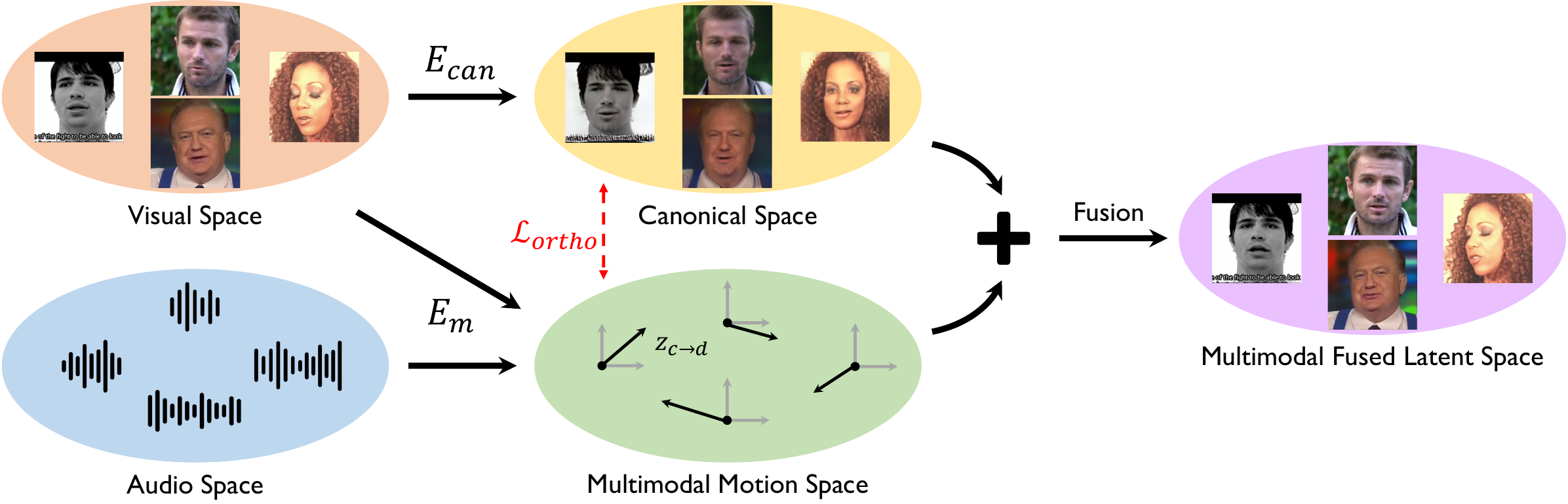}
    \caption{\textbf{Overall framework.} Our goal is to transfer the motions of the individual portrait via latent space exploration. 
    The overall pipeline is comprised of two steps. The first step involves transferring the code in the style latent space to the canonical latent space, where each face has the same facial attributes, but distinct identities.
    In the second step, the multimodal motion space is navigated, where each code contains only motion information and excludes identity information. The target motion code is obtained by fusing image latent features and audio latent features. The canonical code and the motion code are then combined using a linear operation, resulting in the final multimodal fused latent code. This fused latent code is used as input to the decoder to synthesise attribute-controllable talking face videos.
    }
    \vspace{-2mm}
    \label{fig:pipeline}
\end{figure*}

Unlike the previous methods, we grant a higher degree of freedom to the motion vectors rather than restricting them to be constructed by a few orthogonal vectors.
To achieve this goal, we disentangle the latent space of StyleGAN into two distinct subspaces (as shown in ~\Fref{fig:pipeline}): (1) a canonical space that can accommodate a variety of facial identities while maintaining consistent facial attributes, and (2) a multimodal motion space that contains exclusive motion features acquired by fusing both audio and image sources for transferring the target's motion to the source face image.
We enforce that the multimodal motion codes are entirely disentangled from the canonical features by introducing an orthogonality constraint between the canonical space and the multimodal motion space. As a result, the proposed FC-TFG framework manipulates the latent code with a straightforward linear operation, which enables our system to generate more controllable facial animations while avoiding the unwanted entanglement of different features.

Furthermore, our framework is designed as a single encoder-decoder network that utilises StyleGAN's inversion network as the encoder. By taking advantage of this architecture, FC-TFG can effectively disentangle input images into canonical and motion features using simple MLP layers added on top of the inversion network, without significantly increasing the model complexity.

In summary, we make three key contributions:
(1) We propose a novel framework, Fully-Controllable Talking Face Generation (FC-TFG), that generates talking faces with controllable target motion, including head pose, eyebrows, eye blinks, eye gaze, and lip movements.
(2) We separate the style latent space into a canonical space that contains only person-specific characteristics and a multimodal motion space that contains person-agnostic motion features encoded from driving pose video and audio source. 
By imposing an orthogonality constraint on the correlation between the two spaces,  the proposed model produces detailed and controllable facial animation.
(3) We demonstrate that the proposed FC-TFG is highly effective in generating talking faces with sophisticated motion control, producing state-of-the-art results in both qualitative and quantitative metrics. This success highlights the potential of FC-TFG for a wide range of applications that demand elaborate control over various facial features.

\section{Related Works}

\subsection{Audio-Driven Talking Face Generation.}

The synthesis of speech-synchronised video portraits has been a significant area of research in computer vision and graphics for many years~\cite{chen2020comprises,zhu2021deep}. Early works~\cite{fan2015photo,fan2016deep} focus on the individual speaker setting where a single model can generate various talking faces corresponding to a single identity. Recently, the development of deep learning has enabled the design of more generalised talking face generation models~\cite{chung2017you,song2018talking,kr2019towards,zhou2019talking,zhou2020makelttalk,prajwal2020lip,park2022synctalkface} that produce talking faces by inputting identity conditions. However, these works neglect head movements due to the challenge of disentangling head poses from identity-related facial characteristics.

To produce talking face videos that have dynamic and natural movements, there exist studies that utilise landmarks or mesh information ~\cite{chen2019hierarchical,das2020speech,song2022everybody,thies2020neural,yi2020audio}.
For the similar purpose, several works~\cite{deng2019accurate,jiang2019disentangled,bulat2017far,ma2023styletalk,wang2022one} adopt 3D information as intermediate representations. 
Even with additional modalities, the proposed models have certain limitations such as a lack of pose control and the inability to render faces with visually pleasing quality under one-shot conditions. 
Additionally, they suffer from severe performance degradation in situations where the accuracy of landmarks is low, especially in the wild. 

Several recent works~\cite{zhou2021pose,liang2022expressive,min2022styletalker,burkov2020neural, wang2022progressive, hwang2023discohead} have demonstrated that it is possible to create realistic talking faces that mimic different movements and identities from a target video. However, there are a number of limitations: \cite{zhou2021pose} can only modify head pose, \cite{min2022styletalker} cannot animate eye gazes, and \cite{liang2022expressive, wang2022progressive, hwang2023discohead} require facial keypoints to separate visual information for generating detailed expressions.

Unlike the previous works, our framework can generate a wide range of target facial features including pose, lip, eye blink, and even eye gaze without requiring extra annotations or structural information.
Our framework seeks a canonical space of a generator, where each face has the same lip shape and pose but different identities. 
This method enables the creation of more advanced facial representations and simplifies the modeling of motion transfer between source and target images by changing their motion relationship from relative to absolute.
Additionally, our model consistently produces high-quality videos under one-shot conditions.

\subsection{Latent Space Editing.}

Latent space editing involves intentionally modifying the generated output results by exploring meaningful directions in a high-dimensional latent space of a pre-trained generator network. 
These directions enable intuitive navigation that corresponds to desired image manipulation. To manipulate the latent space, some approaches~\cite{shen2020interpreting,jahanian2019steerability,goetschalckx2019ganalyze} directly propagate labeled supervision such as facial attributes. Other works~\cite{voynov2020unsupervised,peebles2020hessian,shen2021closed,yang2021l2m} demonstrate the potential of modifying the semantics of images without annotation.

A recent work~\cite{wang2022latent} achieves success in applying latent space editing technology to face reenactment task which aims to transfer target motion to source image using RGB modality alone.

In contrast to finding directions that correspond to individual facial movements in the latent space by using a single modality, our work aims to disentangle the face identity and the complex attributes composing face motions with multimodal features representing both visual and audio information.
\section{Method}

\begin{figure*}[!t]
    \centering
    \includegraphics[width=0.85\linewidth]{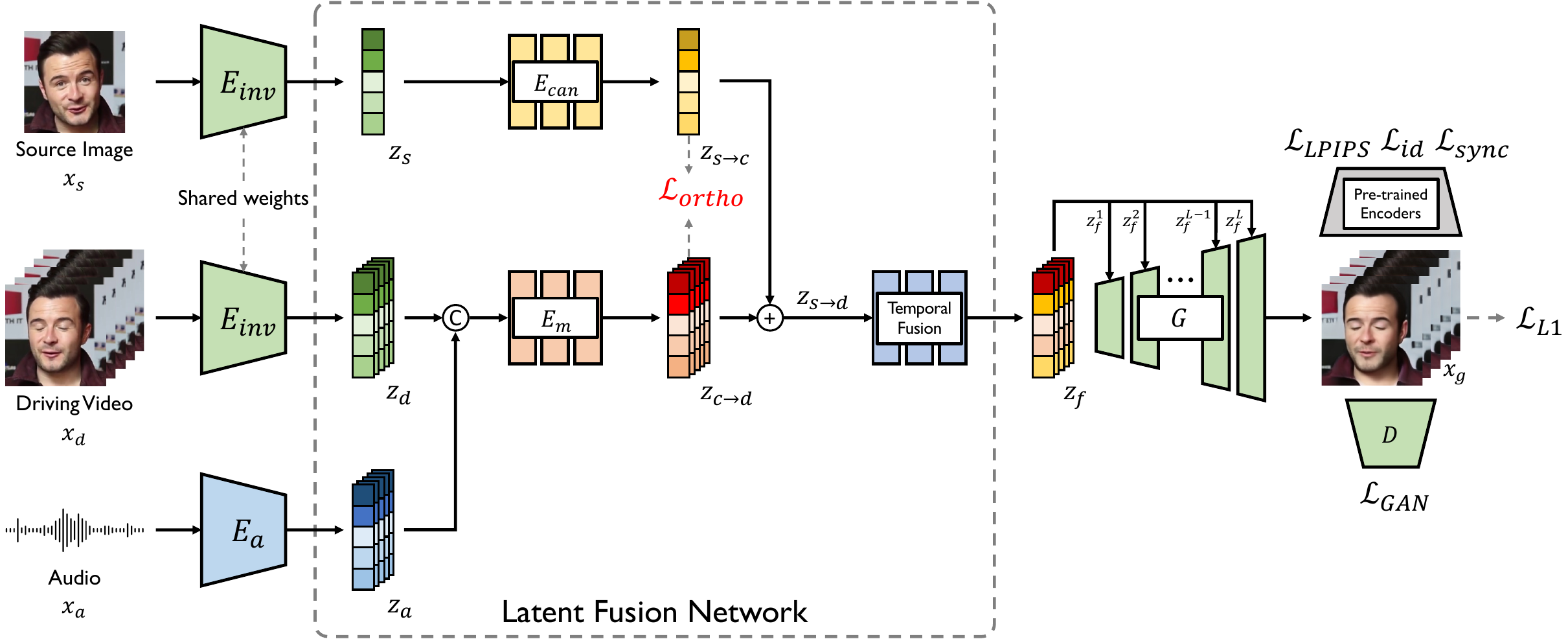}
    \caption{\textbf{Overall Architecture of FC-TFG.} Our model consists of several components, including a visual encoder $E_{inv}$, an audio encoder $E_a$, a generator $G$, a discriminator $D$, and a Latent Fusion Network. For compactness, we use a single visual encoder to extract both a source latent code $z_s$ and a driving latent code $z_d$. The audio encoder extracts an audio latent code $z_a$. First, we design a canonical encoder $E_{can}$ to map $z_s$ to a canonical space as $z_s \rightarrow z_{s \rightarrow c}$, and then linearly combine it with a target motion code $z_{c \rightarrow d}$. This motion code is produced by a multimodal motion encoder $E_m$, which combines $z_d$ and $z_a$. To generate natural motions, we pass the target code through a Temporal Fusion layer before feeding it into the generator $G$. The generated video $x_g$ is compared to a driving video $x_d$ based on its visual and synchronisation quality. Note that we employ the StyleGAN2 generator as a decoder to control coarse-fine motion transfers, and $L$ is the number of modulation layers of the generator. Therefore, each model ($E_{can}$ and $E_m$) in Latent Fusion Network has $L$ independent weights.
    }
    \label{fig:overall}
\end{figure*}

In this work, we propose a self-supervised approach for generating realistic talking face videos by transferring complex motions such as lip movements, head poses, and eye gazes from a driving video to a source identity image. 
Such motion transformation is modeled via disentangled latent feature manipulation.
The proposed framework's pipeline is illustrated in~\Fref{fig:pipeline}.
The key to our framework is two mapping operations: (1) Visual Space to Canonical Space, (2) Visual/Audio Space to Multimodal Motion Space. Through the first mapping, we can obtain canonical images that have the same motion features and different identities. Meanwhile, the second mapping yields motion vectors that enable us to transfer desired motions onto canonical images. To ensure the disentanglement of the two subspaces, we impose an orthogonality constraint. Based on this process, our model is capable of generating natural-looking talking face videos that mimic the full facial motions of the target.

We further provide an overall architecture of the proposed framework in~\Fref{fig:overall}. Our framework includes a visual encoder $E_{inv}$, an audio encoder $E_a$, a generator $G$, a discriminator $D$, and a Latent Fusion Network. We adopt a single shared visual encoder to extract both source and driving latent codes, denoted as $z_s$ and $z_d$, respectively. The audio encoder extracts an additional audio latent code, $z_a$. The canonical encoder $E_{can}$ maps $z_s$ to a canonical code, denoted as $z_{s \rightarrow c}$, which is then combined with a motion code $z_{c \rightarrow d}$ obtained from the multimodal motion encoder $E_m$.
Since an orthogonality constraint is imposed between $z_{s \rightarrow c}$ and $z_{c \rightarrow d}$, the motion transfer process is implemented by simply adding the two latent codes.
To ensure generated motions are natural, the fused code $z_{s \rightarrow d}$ is passed through a Temporal Fusion layer before being fed into the generator $G$. The generator, which is the StyleGAN2 decoder, allows for controlling the coarse-fine motion transfers.
Overall, the proposed method presents a promising approach for generating high-quality talking face videos with controllability.

\subsection{Navigating Canonical Space}
\label{Sec:Navigating_Canonical_Space}
Our main target is to obtain a latent code ${z_{s \rightarrow d{:}t}} {\sim} {\mathcal{Z}} {\in} {\mathbb{R}^{N}}$ that captures the intricate motion transformation from the source image $x_{s}$ to the $t$-th driving frame $x_{d{:}t}$. From now on, we omit the temporal index $t$ for better readability.
Directly seeking $z_{s \rightarrow d}$ in the latent space is an arduous task because the model must be able to capture the subtle relative motion relationship between $x_{s}$ and $x_{d}$ while accurately representing the intricate facial attributes~\cite{wang2022latent}.
To address this challenging issue, we adopt an innovative approach proposed in~\cite{siarohin2019first, siarohin2021motion, wang2021one} by assuming the existence of a canonical image $x_c$ in a canonical space that has unified face-related motions but individual identities. Consequently, we can acquire the target mapping code $z_{s \rightarrow d}$ through a two-stage motion transfer process.

\begin{equation}
    z_{s \rightarrow d} = z_{s \rightarrow c} + z_{c \rightarrow d},
\end{equation}
where $z_{s \rightarrow c}$ and $z_{c \rightarrow d}$ indicate the transformation $x_{s} \rightarrow x_{c}$ and $x_{c} \rightarrow x_{d}$ respectively. Now, the mapping $z_{s \rightarrow d}$ can be solved with an absolute motion transfer process, where both $z_{s \rightarrow c}$ and $z_{s \rightarrow d}$ do not need to account for the motion relationship between them.

Unlike the aforementioned previous works~\cite{siarohin2019first, siarohin2021motion, wang2021one}, which rely solely on the RGB modality to discover the canonical space, our method leverages both RGB and audio modalities and independently control the facial motions and lip movements simultaneously.

To design this motion transfer process effectively and intuitively, we apply a simple linear operation that adds $z_{c \rightarrow d}$ directly to $z_{s \rightarrow c}$. Here, $z_{s \rightarrow c}$ is obtained by passing $z_s = E_{inv}(x_s)$ through $E_{can}$ which is a 2-layer MLP, and $z_{c \rightarrow d}$ is learned by navigating motion space that will be further explained in the following section.

\subsection{Navigating Multimodal Motion Space}
\label{Sec:Navigating_Multimodal_Motion_Space}

In order to generate natural head poses, lip motions, and facial expressions simultaneously, $z_{c \rightarrow d}$ should successfully combine both vision and audio information while only containing motion-related features.
To disentangle identity-related codes and motion-related codes in $N$-dimensional latent space,
we inject a channel-wise orthogonality constraint into each channel in $z_{s \rightarrow c}$ and $z_{c \rightarrow d}$.

By strictly enforcing this constraint, the model induces that the identity and motion codes do not interfere with each other when they are combined for the transformation with the linear operation as explained in~\Sref{Sec:Navigating_Canonical_Space}.
Finally, $z_{c \rightarrow d}$ can be obtained as follows:

\begin{equation}
    z_{c \rightarrow d} = E_m(E_a(x_a) \oplus E_{inv}(x_d)),
\end{equation}
where $E_a$, $E_m$, and $x_a$ denote the audio encoder, motion encoder, and audio input respectively. $\oplus$ operation indicates channel-wise concatenation. Note that as we solely traverse the latent space of the pre-trained image decoder, the resulting $z_{s \rightarrow d}$ can retain semantically meaningful features. The structure of the motion encoder is a 3-layer MLP. Finally, to generate temporally consistent latent features, we refine the acquired $z_{s \rightarrow d}$ by feeding it to the Temporal Fusion layer, which consists of a single 1D convolutional layer.

\begin{table*}[!t]
    \centering
    \resizebox{0.95\linewidth}{!}
    {
        \renewcommand{\arraystretch}{1.3}
        \begin{tabular}{l|cccc|ccccc|ccccc}
        \hline\hline
         & \multicolumn{4}{c}{Controllable Motions} & \multicolumn{5}{|c}{VoxCeleb2}  & \multicolumn{5}{|c}{MEAD} \\ 
        \cline{2-15}
        Method & lip & pose & eye blink & eye gaze & {SSIM}$\uparrow$\quad & {MS-SSIM}$\uparrow$\quad & {PSNR}$\uparrow$\quad & {LMD}$\downarrow$\quad & {LSE-C}$\uparrow$\quad & {SSIM}$\uparrow$\quad & {MS-SSIM}$\uparrow$\quad & {PSNR}$\uparrow$\quad & {LMD}$\downarrow$\quad & {LSE-C}$\uparrow$\quad \\ 
        \hline
        Wav2Lip~\cite{prajwal2020lip} & \cmark & \xmark & \xmark & \xmark & 0.58 &  -  & 20.63 & 2.65 & \textbf{8.66} & \textbf{0.85} &  -  & 26.15 & 3.11 & \textbf{7.25} \\      
        MakeItTalk~\cite{zhou2020makelttalk} & \cmark & \xmark & \xmark & \xmark & 0.55 & 0.45 & 16.94 & 3.28 & 3.83 & 0.78 &  0.78  & 23.6 & 3.55 & 4.35 \\
        Audio2Head~\cite{wang2021audio2head} & \cmark & \xmark & \xmark & \xmark & 0.51 & 0.40 & 15.98 & 3.58 & 5.79 & 0.67 &  0.59  & 19.57 & 4.85 & 5.38 \\
        PC-AVS~\cite{zhou2021pose}    & \cmark & \cmark & \xmark & \xmark & 0.57 & 0.60 & 17.37 & 2.25 & 5.82 & 0.66 &  0.7  & 19.95 & 2.93 & 5.19 \\ 
        \hline
        FC-TFG (Ours) & \cmark & \cmark & \cmark & \cmark & \textbf{0.69} & \textbf{0.77} & \textbf{21.22} & \textbf{1.58} & 8.46 & 0.84 &  \textbf{0.89}  & \textbf{26.19} & \textbf{2.46} & 5.51 \\
        \hline\hline
        \end{tabular}
    }
\caption{\textbf{Quantitative Results.} We compare our method to four publicly available baselines on six different metrics. Our approach outperforms the baseline methods in terms of both visual quality and lip synchronisation, while simultaneously controlling diverse and detailed facial motions. We evaluate the generated samples using the original authors' experimental settings, ensuring a fair comparison between the different methods.}
\vspace{-3mm}
\label{tab:main_results}
\end{table*}

\subsection{Training Objectives}

\subsubsection{Orthogonality loss.}
To ensure effective disentangling of $z_{s \rightarrow c}$ and $z_{c \rightarrow d}$, we extend the application of the orthogonality loss~\cite{yang2021l2m} to each layer of the StyleGAN decoder while combining both audio and video information. This approach enables us to manipulate finer movements, enhancing the overall capability of our framework.
The orthogonality loss function is expressed as follows:

\begin{equation}
    \mathcal{L}_{ortho} = {{1}\over{N}} \sum (z_{s \rightarrow c} \astrosun z_{c \rightarrow d}),
\label{eq:ortho}
\end{equation}
where $N$ denotes the number of layers in decoder and $\astrosun$ indicates Hadamard product operation.

\subsubsection{Synchronisation loss.}
To generate well-synchronised videos $x_g$ according to the input audio conditions, we adopt a sync loss function, which leverages a pre-trained SyncNet~\cite{chung2017out} comprising an audio encoder and a video encoder. Many works~\cite{min2022styletalker, park2022synctalkface, prajwal2020lip, wang2022one, ma2023styletalk} modify the SyncNet with altered objective functions, to further improve lip synchronisation quality. We use the modified SyncNet introduced in~\cite{min2022styletalker} to enhance our model's lip representations.

The distance between the features of a video and its synchronised audio, extracted from the pre-trained Syncnet model, should be close to 0. From this, we minimise the following sync loss:
\begin{equation}
    \mathcal{L}_{s y n c}=1-\cos \left(f_v\left(x_g\right), f_a\left(x_a\right)\right),
\end{equation}
where $f_a$ and $f_v$ denote the audio encoder and video encoder of SyncNet respectively.

\subsubsection{Identity loss.}
To preserve facial identity after motion transformation, we apply an identity-based similarity loss~\cite{richardson2021encoding} by employing a pre-trained face recognition network $E_{id}$~\cite{deng2019arcface}.

\begin{equation}
    \mathcal{L}_{id} = 1-\cos \left(E_{id}\left(x_{g}\right), E_{id}\left(x_d\right)\right).
\end{equation}

\subsubsection{Reconstruction loss.}
For the reconstruction loss, we adopt $L1$ loss function that calculates the pixel-wise $L1$ distance between the generated talking face image, $x_g$, and the target image, $x_d$. The reconstruction loss can be calculated as follows:

\begin{equation}
    \mathcal{L}_{rec} = \parallel x_g - x_d \parallel_1.
\end{equation}

\subsubsection{Perceptual loss.}
Using $L1$ reconstruction loss alone may result in blurry images or slight artifacts as it is a pixel-level loss. To compensate for the smoothing effect caused by the reconstruction loss $\mathcal{L}_{L1}$, we add the Learned Perceptual Image Patch Similarity (LPIPS) loss~\cite{zhang2018unreasonable}, which measures the perceptual similarity between two images. LPIPS loss can be calculated as follows:

\begin{equation}
    \mathcal{L}_{LPIPS} = \frac{1}{N_f}\sum_{i=1}^{N_f} \parallel \phi(x_g)_i - \phi(x_{d})_i \parallel_2,
\end{equation}
where $\phi$ is a pre-trained VGG19~\cite{simonyan2014very} network, and $N_f$ is the number of feature maps.

\subsubsection{Adversarial loss.}
We perform adversarial training with an image discriminator, $D$, to improve the quality of the generated videos. The architecture of $D$ is the same as StyleGAN2~\cite{karras2020analyzing} discriminator. We use a non-saturating loss~\cite{goodfellow2020generative} for adversarial training, which can be expressed as follows:
\begin{equation}
\begin{split}
    \mathcal{L}_{GAN} = \min_{G}\max_{D}\Bigl(&\mathbb{E}_{x_d}[\log(D(x_d))] \\
    &+ \mathbb{E}_{z_f}[\log(1-D(G(z_f))] \Bigr).
\end{split}
\end{equation}

\subsubsection{Overall loss.}
Our total loss can be formulated as follows:

\begin{equation}
\begin{split}
    \mathcal{L}_{total} = \lambda_{1}\mathcal{L}_{ortho} + \lambda_{2}\mathcal{L}_{sync} + \lambda_{3}\mathcal{L}_{id} \\ + \lambda_{4}\mathcal{L}_{rec} + \lambda_{5}\mathcal{L}_{LPIPS} + \lambda_{6}\mathcal{L}_{GAN},
\end{split}
\end{equation}
where hyperparameters $\lambda$ are introduced to balance the scale of each loss. Each $\lambda$ controls the relative importance of its corresponding loss term. Empirically, $\lambda_{1}$, $\lambda_{2}$, $\lambda_{3}$, $\lambda_{4}$, $\lambda_{5}$, and $\lambda_{6}$ are set to 1, 0.1, 0.5, 1, 1 and 0.1 respectively.

\section{Experiments}

\subsection{Experimental Setup}
\subsubsection{Dataset.}
Our framework is trained on VoxCeleb2~\cite{chung2018voxceleb2} and evaluated on both VoxCeleb2 and MEAD~\cite{wang2020mead}.
VoxCeleb2 includes 6,112 different identities and over 1 million utterances. Of the total identities, 5,994 are used for training, while the remaining identities are reserved for testing. We follow the pre-processing procedure proposed in~\cite{siarohin2019first} to ensure the consistent visual quality of videos.

To evaluate one-shot talking face generation performance, we utilise the MEAD dataset, which comprises emotional faces featuring more than 30 actors and eight emotion categories at three intensity levels. To conduct the evaluation, we randomly choose five speakers and five videos for each emotion category, and we only use the frontal-view videos from this dataset for testing purposes.

\subsubsection{Implementation Details.}
First of all, we pre-train a StyleGAN2~\cite{karras2020analyzing} generator on the VoxCeleb2 dataset and then train HyperStyle~\cite{alaluf2022hyperstyle} inversion network with the pre-trained StyleGAN2 model. Specifically, we replace the e4e~\cite{tov2021designing} encoder in the HyperStyle model with pSp~\cite{richardson2021encoding} encoder.
We focus on manipulating 8 specific layers of 14 layers in our generator, namely layers 1, 2, 3, 4, 7, 8, 9, and 10. Additionally, we only input the audio feature into 2 specific layers, layers 7 and 8. This allows us to effectively control the style of the generated images based on the audio input.

For the audio source, we downsample the audio to 16kHz, then convert the downsampled audio to mel-spectrograms with a window size of 800, a hop length of 200, and 80 Mel filter banks. We utilise a pre-trained audio encoder 
 introduced in~\cite{prajwal2020lip}. We use Adam~\cite{kingma2014adam} optimiser for updating our model, with a learning rate of ${1e{-4}}$.
Our framework is implemented on PyTorch~\cite{paszke2019pytorch} and trained with eight 48GB A6000 GPUs. Note that the aforementioned pre-trained models are fine-tuned during the training stage.

\subsubsection{Comparison Methods.}
We compare our method with several state-of-the-art talking face synthesis methods, which are publicly available. \textbf{Wav2Lip}~\cite{prajwal2020lip} employs pre-trained SyncNet as a lip-sync discriminator to generate well-synchronised mouth region of the source image. \textbf{MakeItTalk}~\cite{zhou2020makelttalk} predicts landmarks through 3D face models and generates both lip movements and head motions simultaneously driven by audio. \textbf{Audio2Head}~\cite{wang2021audio2head} generates head motions by utilising a keypoint-based dense motion field driven by audio. \textbf{PC-AVS}~\cite{zhou2021pose} is a pose-controllable talking face generation model that controls head poses with driving videos.

\begin{figure*}[!t]
    \centering
    {
    \includegraphics[width=0.92\linewidth]{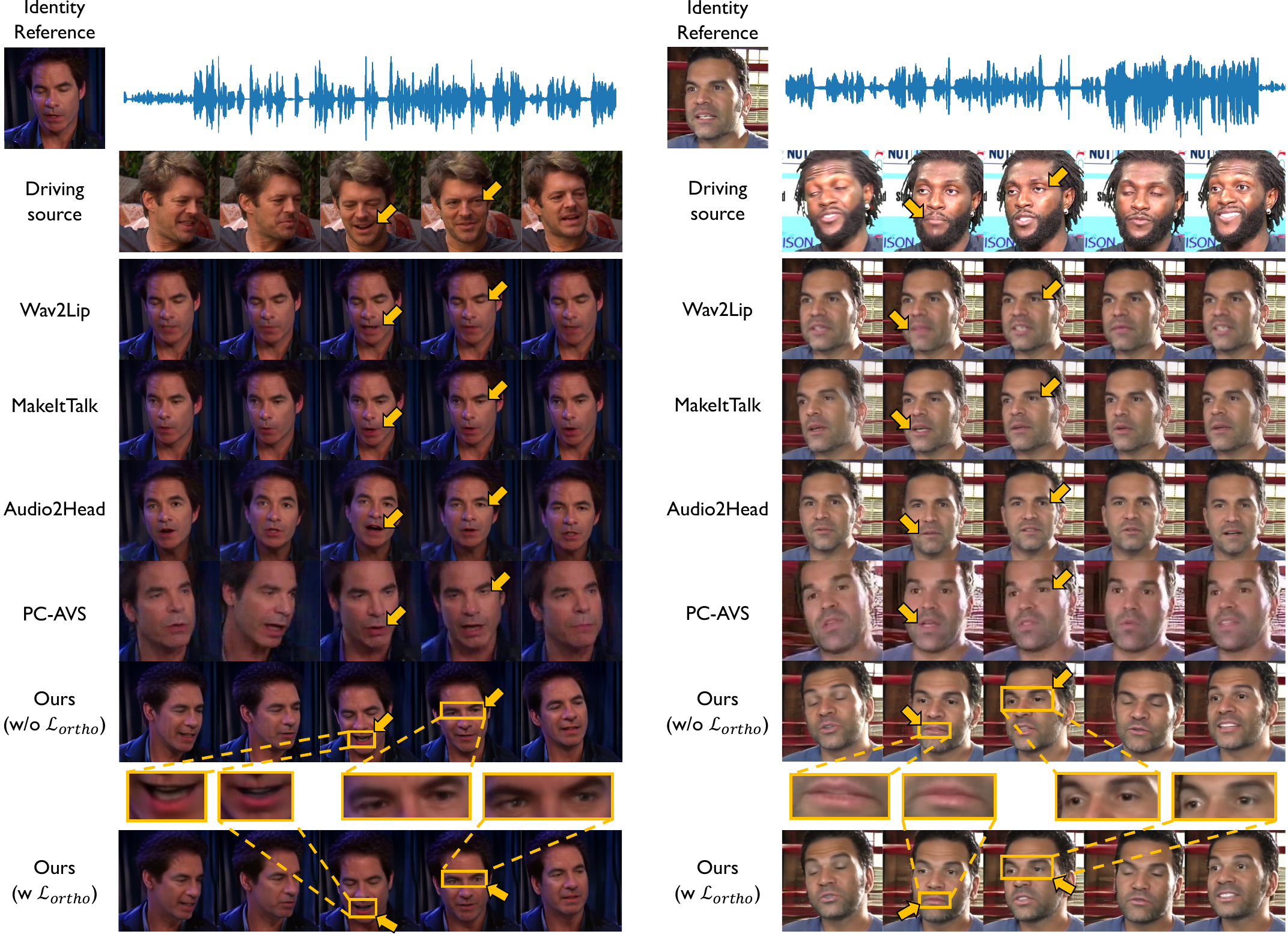}
    }
    \vspace{-2mm}
    \caption{{\textbf{Qualitative Results.} We compare our method with several baselines listed in~\Tref{tab:main_results}. Our approach outperforms all the baselines in terms of generating precise head motion and facial expressions following the given conditions. Wave2Lip, MakeItTalk, and Audio2Head fail to generate accurate head motion of the driving source videos. PC-AVS produces a similar head pose with target motion but lacks in generating realistic facial expressions. On the other hand, our method successfully generates every facial expression of the target motion while synchronising the lip with the input audio source.}}
    \label{fig:qualitative}    
\end{figure*}

\subsection{Quantitative Results}

\subsubsection{Evaluation Metrics.}
We conduct quantitative evaluations with various evaluation metrics that have previously been adopted in the talking face generation field. To account for the accuracy of mouth shapes and lip sync, we use Landmarks Distance (LMD) around the mouths proposed in~\cite{chen2019hierarchical}, and Lip Sync Error Confidence (LSE-C) proposed in~\cite{chung2017out}. To compare the visual quality of the generated videos, we calculate Structural Similarity Index Measure (SSIM)~\cite{wang2004image}, Multi Scale Structural Similarity Index Measure (MS-SSIM)~\cite{wang2003multiscale}, and Peak Signal-to-Noise Ratio (PSNR).

\subsubsection{Controllable Motion Types.}
In order to assess the capabilities of the baseline methods in generating realistic talking faces, we compare the number of controllable motions for each method under the target conditions in Table~\Tref{tab:main_results}. Wav2Lip, MakeItTalk, and Audio2Head rely solely on audio input to generate videos, resulting in limited control over motion and pose. In particular, MakeItTalk, and Audio2Head are capable of generating lip movements and small head pose variations that are synchronised with the audio conditions but are unable to produce more complex and diverse motion patterns. On the other hand, PC-AVS and Ours have the capacity to control head pose following the target pose conditions. However, while PC-AVS is limited to controlling head pose, our approach is effective for controlling all facial attributes, including eye blinks and gazes, by exploring disentangled latent spaces. 
\begin{table*}[!t]
    \centering
    \resizebox{0.73\linewidth}{!}
    {
        \renewcommand{\arraystretch}{1.4}
        \begin{tabular}{l|ccccc} 
        \hline\hline
        Method & \multicolumn{1}{|c}{Wav2Lip} & \multicolumn{1}{c}{MakeItTalk} & \multicolumn{1}{c}{Audio2Head} & \multicolumn{1}{c}{PC-AVS} & \multicolumn{1}{c}{\textbf{FC-TFG (Ours)}}\quad \\ 
        \hline
        Lip Sync Quality$\uparrow$            & 3.47 & 2.31 & 2.55 & 3.29 & \textbf{3.93}\\
        Head Movement Naturalness$\uparrow$   & 1.88 & 2.5 & 2.98 & 3.23 & \textbf{4.14}\\
        Overall Quality$\uparrow$              & 2.18 & 2.64 & 2.91 & 2.91 & \textbf{3.94}\\
        \hline\hline
        \end{tabular}
    }
\caption{\textbf{User Study.} We conduct a user study on generated videos with three aspects: lip synchronisation, the naturalness of head movement, and overall video quality. The higher the better, with the value range of 1 to 5.}
\vspace{-3mm}
\label{tab:mos}
\end{table*}

\subsubsection{Evaluation Results.}
We follow the evaluation protocol introduced in~\cite{zhou2021pose}. Specifically, we select the first frame of each test video to serve as a reference identity. We use the remaining frames to determine the subject's pose, facial expression, and lip shape. 

As shown in~\Tref{tab:main_results}, we compare our method with four baselines on VoxCeleb2 and MEAD datasets.
For VoxCeleb2 dataset, except for LSE-C metric, our proposed method surpasses the previous audio-driven and pose-controllable methods in all metrics while manipulating detailed facial components. 
For MEAD dataset, our framework shows the best performance on MS-SSIM, PSNR, and LMD and comparable results to the Wav2Lip model in terms of SSIM. 
Although the Wav2Lip model shows better performance on LSE-C metric, the generated face shows significant distortions.
These results demonstrate that our model is capable to capture detailed facial movements and generate visually pleasing results by keeping semantically meaningful the latent space.

\subsection{Qualitative Results}

\subsubsection{Fully-Controllable Talking Face Generation.} We visually show our qualitative results in~\Fref{fig:qualitative}. We would like to clarify that Wav2Lip is a model that targets to change only the lip region. MakeItTalk and Audio2Head are models that generate natural face motions conditioned by only audio sources. For that reason, the generated videos cannot replicate diverse facial motions in the driving videos. PC-AVS focuses on synthesising the talking faces with controllable head pose variations. However, it fails to mimic the diving source's facial expressions such as eye blinks, eyebrows movements, and eye gazes. On the other hand, our method can generate fully-controllable talking faces by synthesising both head pose and facial expressions that precisely follow the driving source videos. 

In the left example of~\Fref{fig:qualitative}, our model captures the driving source's smiles and eye gaze, as indicated by the yellow arrows. Similarly, in the right example, our model replicates detailed facial movements such as lip shape, eyebrow position, forehead frowns, and eye gazes. 
This high level of accuracy and detail allows our method to generate realistic and expressive talking faces that closely mimic the emotions and expressions of the driving source.

\begin{figure}[!t]
    \centering
    \includegraphics[width=0.95\linewidth]{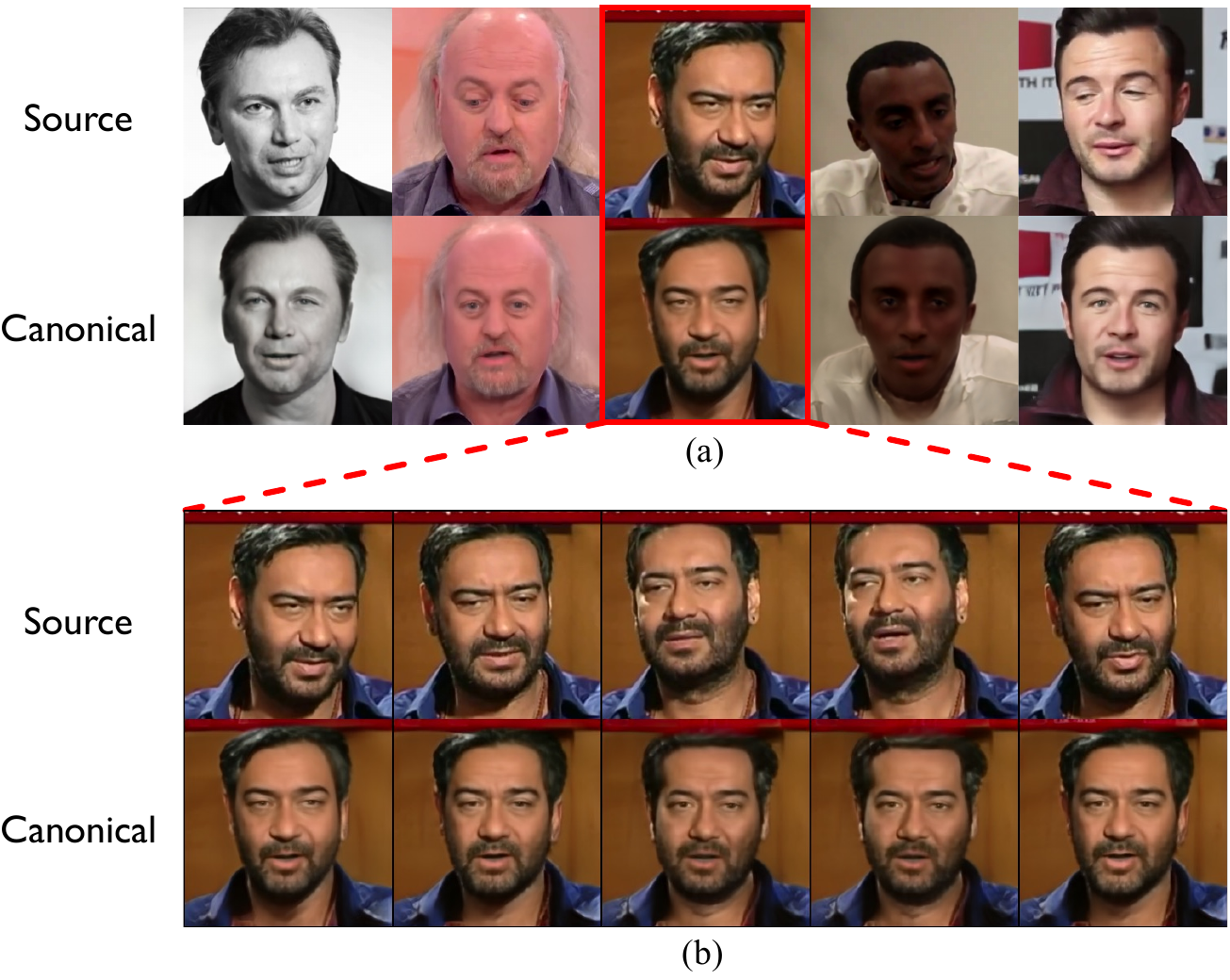}
    \caption{\textbf{Samples in Canonical Space.} We demonstrate how well our model preserves identity by mapping various identities to the canonical space. In Fig. (a), we generate diverse canonical image samples having different identities by feeding each canonical code to our generator. In Fig. (b), we further visualise every canonical image from a single video. These results prove that our model is robust to maintain the source identities and well-generalised to various identities.
    }
    \label{fig:canonical}
\end{figure}

\subsubsection{User Study.}
We assess the quality of the videos generated by FC-TFG with a user study of 40 participants for their opinions on 20 videos. Specifically, we randomly collect reference images, driving videos, and driving audios from VoxCeleb2~\cite{chung2018voxceleb2} test split. Subsequently, we create videos based on Wav2Lip~\cite{prajwal2020lip}, MakeItTalk~\cite{zhou2020makelttalk}, Audio2Head~\cite{wang2021audio2head}, PC-AVS~\cite{zhou2021pose}, and FC-TFG (Ours). We adopt the widely used Mean Opinion Scores (MOS) as an evaluation metric following~\cite{liang2022expressive,zhou2021pose}. Each user gives evaluation scores from 1 to 5 for the following aspects: (1) lip sync quality; (2) head movement naturalness; and (3) video realness. The order of methods within each video clip is randomly shuffled. Note that for a fair comparison with fully audio-driven generation methods, we do not evaluate how well the generated head poses are matched to the poses of the driving videos. 
As shown in~\Tref{tab:mos}, it is demonstrated that our method generates talking face videos with higher lip synchronisation and natural head movement, compared to the existing methods.

\subsection{Ablation Study}

\subsubsection{Canonical Space.}
In~\Fref{fig:canonical} (a), we present the impressive outcomes of mapping various identities to the canonical space \emph{where each face has the same facial motion but different identities}. The effectiveness of our approach in preserving identity can be clearly observed through the distinct and recognisable facial features of each individual. Additionally, as shown in~\Fref{fig:canonical} (b), regardless of the differences in the input images having the same identity, our approach consistently produces similar canonical images that accurately capture the individual's distinct facial identity and detailed features. These results further highlight the effectiveness of our method in finding the canonical space that is essential for transferring complex facial motions with a simple linear operation. 

\begin{table}[h]
    \centering
    \resizebox{1.0\linewidth}{!}
    {
        \centering
        \renewcommand{\arraystretch}{1.35}
        \begin{tabular}{cc|cc|ccccc}
        \hline\hline
        \multicolumn{2}{c|}{Temporal Fusion} & \multicolumn{2}{c|}{$D$} & \multicolumn{5}{c}{Metrics}                \\ 
        \hline
        1D-Conv & LSTM          & image & video        & SSIM$\uparrow$ & MS-SSIM$\uparrow$ & PSNR$\uparrow$ & LMD$\downarrow$ & LSE-C$\uparrow$ \\ 
        \hline
         & \cmark & \cmark &  & 0.66 & 0.75 & 20.78 & 1.79 & \textbf{8.66} \\ 
         & \cmark &  & \cmark & 0.67 & 0.75 & 20.64 & 1.73 & 7.9 \\ 
        \cmark &  &  & \cmark & 0.66 & 0.75 & 20.72 & 1.98 & 8.04 \\ 
        \cmark &  & \cmark &  & \textbf{0.69} & \textbf{0.77} & \textbf{21.22} & \textbf{1.58} & 8.46     \\ 
        \hline\hline
        \end{tabular}


    }
\caption{\textbf{Ablations on Model Design.} We report performance results based on our different model choices. We observe that the highest performance is achieved when using a 1D convolutional layer for the Temporal Fusion layer and an image discriminator for the discriminator.}
\vspace{-5mm}
\label{tab:design}
\end{table}

\subsubsection{Ablations on Model Design.}
We conduct ablation studies on our model choices for the Temporal Fusion layer and Discriminator $D$ in~\Tref{tab:design}.
We empirically observe that a flickering phenomenon appears when using Long Short-Term Memory (LSTM) network as our temporal fusion model. We suspect that this issue is caused by our window size, which does not have any overlap along the temporal dimension for faster inference. We solve the flickering issue by replacing the LSTM layer with a 1D convolutional layer.

We investigate two different types of discriminators in this work; (1) image discriminator, and (2) video discriminator. The former is designed to evaluate the authenticity of a single image input, while the latter is trained to classify a sequence of frames concatenated along the channel dimension as either real or fake video.

As shown in~\Tref{tab:design}, the lip sync confidence score is high when we use the image discriminator. We believe that this is because the image discriminator is designed to analyse the details of a single image, which means that the generator must focus on spatial information to deceive the discriminator. On the other hand, the video discriminator examines the authenticity of multiple frames at the same time, which requires the generator to focus on both spatial and temporal information. Based on our analysis, we conclude that the image discriminator is more suitable for our intended purpose of accurately generating lip shape, which is a spatially small region.

\subsubsection{Orthogonality Constraint Effectiveness.}
\begin{figure}[!t]
    \centering
    \includegraphics[width=0.85\linewidth]{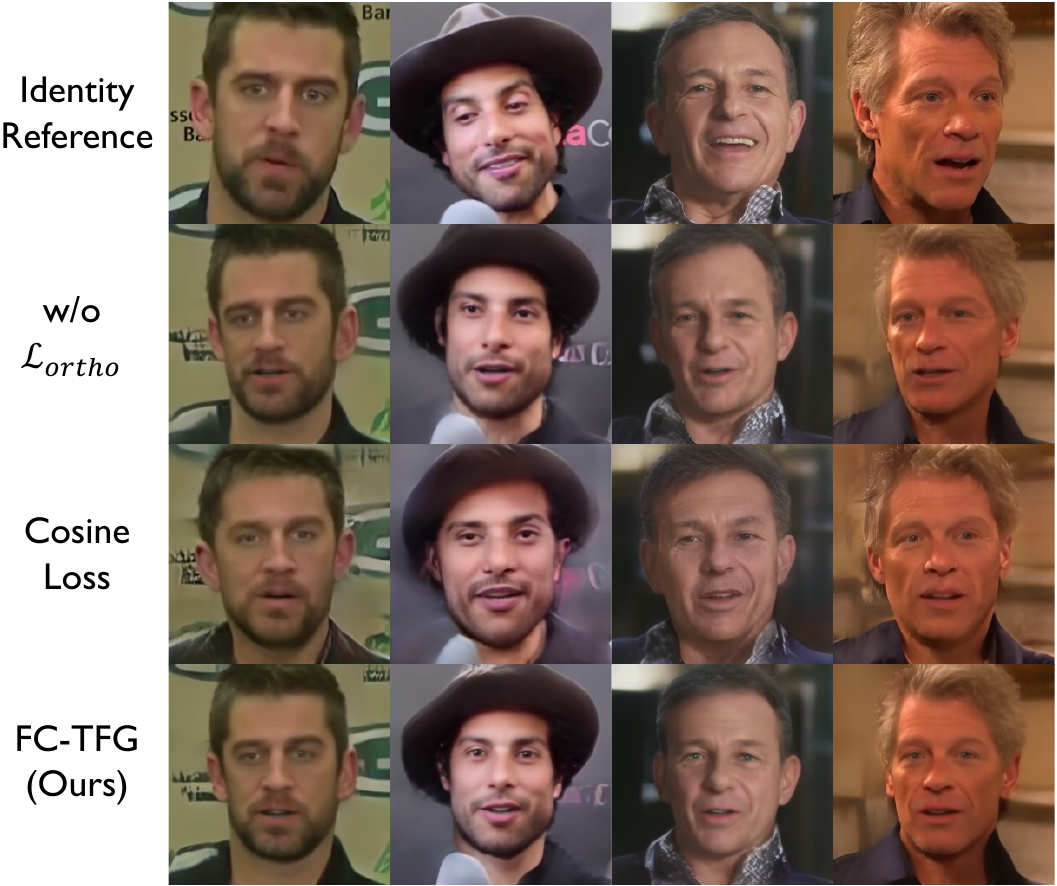}
    \caption{\textbf{Analysis on the effectiveness of Orthogonality Constraint.} Our model effectively disentangles the identity and motion information by enforcing orthogonality between the canonical space and the multimodal motion space. This is demonstrated by the cohesive and consistent motion patterns in the canonical space, which is not achieved in the models either trained without the orthogonality constraint or trained with other kinds of constraint such as cosine loss. 
    } 
    \label{fig:orthogonal}
    \vspace{1mm}
\end{figure}

We provide further insight into the efficacy of the orthogonality constraint by visually analysing samples in the canonical space. As depicted in~\Fref{fig:orthogonal}, the canonical space of the model trained without orthogonality constraint contains diverse head poses, indicating the potential lack of disentanglement between identity and motion information. Additionally, when cosine similarity loss is used as the orthogonality constraint, the model produces even more blurry images.
On the contrary, the canonical space of FC-TFG trained with $L_{ortho}$ in~\Eref{eq:ortho} contains more unified facial motions, indicating a successful disentanglement of identity and motion information in the latent space. The quantitative results according to the different orthogonality constraints are reported in~\Tref{tab:ortho}. These results strongly suggest that the orthogonality constraint plays a crucial role in achieving high-quality outcomes.

\section{Conclusion}
In this work, we propose a framework named Fully-Controllable Talking Face Generation (FC-TFG), which can generate every facial expression while synchronising lip movements with the input audio sources. Our framework is carefully designed to disentangle the latent space of StyleGAN into the canonical space and the multimodal motion space. This disentanglement enables the generation of more detailed and controllable facial animations while avoiding the unwanted mixing of different types of conditions. With various experiments, we prove that the proposed method is highly effective, achieving state-of-the-art results in both qualitative and quantitative evaluations of generated video quality. The potential of FC-TFG for a wide range of applications that demand precise control over various facial features is significant, such as virtual reality, and augmented reality. It could also be useful for creating personalised digital avatars or virtual assistants that can communicate and interact with users in a more natural and realistic manner.

\begin{table}[!t]
    \centering
    \resizebox{0.85\linewidth}{!}
    {
        \centering
        \renewcommand{\arraystretch}{1.1}
        \begin{tabular}{c|ccccc}
        \hline\hline
        \multicolumn{1}{c|}{Orthogonal} & \multicolumn{5}{c}{Metrics} \\ 
        \cline{2-6}
        Constraint        & SSIM$\uparrow$ & MS-SSIM$\uparrow$ & PSNR$\uparrow$ & LMD$\downarrow$ & LSE-C$\uparrow$ \\ 
        \hline
        \xmark & 0.67 & 0.76 & 20.88 & 1.70 & 8.27 \\
        Cosine & 0.67 & 0.75 & 21 & 2.11 & 8.38 \\
        $L_{ortho}$ & \textbf{0.69} & \textbf{0.77} & \textbf{21.22} & \textbf{1.58} & \textbf{8.46} \\
        \hline\hline
        \end{tabular}
    }
\caption{Ablations on Orthogonality Loss. We ablate the effectiveness of the orthogonality constraint. The model trained with the proposed loss $L_{ortho}$ shows the best performance.}
\label{tab:ortho}
\end{table}



\clearpage


\bibliographystyle{ACM-Reference-Format}
\bibliography{shortstrings, sample-base}










\end{document}